\documentclass[letterpaper, 10 pt, conference]{ieeeconf}  

\IEEEoverridecommandlockouts                              

\usepackage{graphicx} 

\usepackage{booktabs}

\title{\LARGE \bf
VPAIR - Aerial Visual Place Recognition and Localization \\ in Large-scale Outdoor Environments
}

\author{Michael Schleiss$^{1,2}$, Fahmi Rouatbi$^{1}$ and Daniel Cremers$^{2}$
\thanks{$^{1}$Department Sensor Data Fusion, Fraunhofer FKIE, 53343 Wachtberg, Germany }%
\thanks{$^{2}$Department of Informatics, Technical University of Munich, 85748 Garching, Germany }%
\thanks{Corresponding author: { michael.schleiss@fkie.fraunhofer.de}}
}

\begin{document}

\maketitle
\thispagestyle{empty}
\pagestyle{empty}

\begin{abstract}
Visual Place Recognition and Visual Localization are essential 
components in navigation and mapping for autonomous vehicles especially 
in GNSS-denied navigation scenarios. Recent work has focused on ground or close to ground 
applications such as self-driving cars or indoor-scenarios and low-altitude 
drone flights. However, applications such as Urban Air Mobility require 
operations in large-scale outdoor environments at medium to high altitudes. 
We present a new dataset named VPAIR. The dataset was recorded on board a light 
aircraft flying at an altitude of more than 300 meters above ground capturing
images with a downwardfacing camera. Each image is paired with a high resolution 
reference render including dense depth information and 6-DoF reference poses. 
The dataset covers a more than one hundred kilometers long trajectory over 
various types of challenging landscapes, e.g. urban, farmland and 
forests. Experiments on this dataset illustrate the
challenges introduced by the change in perspective to a bird's eye view 
such as in-plane rotations. The dataset will be made publicly available under https://github.com/AerVisLoc/vpair. 
\end{abstract}

\section{Introduction}

Following the increased interest in autonomous vehicles research a number 
of datasets have been released that address Visual Place Recognition (VPR)
and/or Visual Localization (VL) from the perspective of the self-driving car, 
e.g. \cite{warburg2020mapillary, RobotCarDatasetIJRR}. 
Other datasets were recorded from modalities such as ground robots 
\cite{leyva2019tb}, a railroad system \cite{olid2018single}
handheld-devices \cite{taira2018inloc}
or aerial platforms in indoor \cite{Burri25012016} or  
low-altitude-outdoor environments \cite{maffra2018tolerant}. 
However, they all share a similar camera 
perspective where the scene is close to the camera's position and the 
camera is oriented mostly in an upright manner. 

On the other hand there is an increasing number of applications for 
autonomous flying vehicles in large-scale outdoor scenarios e.g. logistics
\cite{scott2017drone}, patrolling \cite{girard2004border}, inspection
\cite{nikolic2013uav}
and even personal transportation \cite{planing_pinar_2019}. 
As regulatory bodies start to plan to integrate 
autonomous unmanned aerial vehicles into regular airspace by the end of the 
decade \cite{undertaking2018european} there
will be a need for alternatives to satellite-based navigation.  One could 
argue that at high altitudes the reception quality of satellite signals 
would be favorable. However, global satellite navigation systems (GNSS) are
very vulnerable to accidental or malicious radio frequency interruptions 
(jamming) or fake signals (spoofing) \cite{9340116}. Safe aviation requires 
back up systems, especially without a human pilot. True autonomous flight 
thus can only become a reality at large scale if back up systems for 
GNSS-failure exist. 

In VPR it is the goal to retrieve a coarse camera pose for a given image by 
querying a large database of geotagged images for instances 
of the same place. This can then be followed by a VL-based pose refinement step
in a coarse to fine manner \cite{sarlin2019coarse}. Both techniques together 
provide the capability to retrieve a drift-free 6-DoF global position estimate
in absence of GNSS but require a database with precise reference imagery. 
The database is often built using traversals of the 
same place at different times. Similar to \cite{vallone2022danish}, we 
instead use publicly available geodata to render reference imagery and 
dense depth maps. 

Matching query and reference imagery is challenging due to variation in appearance 
induced by illumination, weather and seasons, as well as viewpoint variations.
Learned global and local image descriptors such as NetVLAD 
\cite{arandjelovic2016netvlad} and D2-Net \cite{dusmanu2019d2} have shown promising performance in the context
of large-scale autonomous navigation scenarios \cite{zaffar2021vpr,sattler2018benchmarking}. 
Our experiments, however show, that the performance degrades drastically when faced with in-plane rotations, 
which is typical for an aerial setting with a downward-facing camera. 
Based on recent work by Parihar et al. \cite{parihar2021rord} we show how
rotation-robust features provide promising results both for VPR and VL 
highlighting the need to adapt current state-of-the-art 
techniques to the aerial scenario.

By releasing our dataset we hope to support the aerial robotics community in 
developing and extending VPR and VL techniques to the aerial scenario 
to enable safe autonomous flight in large-scale environments.


\section{Related Work}

VPR and VL datasets usually follow a 
structure where a set of query images that capture the view of a device or a 
vehicle are accompanied by a set of geotagged reference images from 
one or multiple temporally separate traversals of the same spatial locations.
Based on their geotags the reference images provide a coarse position estimate.
These can subsequently be used as a starting point for a pose refinement step.

Based on the modality from which the query and reference images were captured these 
datasets can be categorized into ground-to-ground, aerial-to-aerial or aerial-to-ground 
datasets. Ground-to-ground datasets such as \cite{torii2013visual, warburg2020mapillary,leyva2019tb,olid2018single, taira2018inloc} 
use the same sensor setup to collect query and reference images. 
The same is true for aerial-to-aerial VPR datasets such as \cite{Burri25012016, maffra2018tolerant}.
More recently aerial-to-ground datasets have introduced the idea of using 
external geodata for geolocating ground vehicles from aerial imagery such as orthophotos.
External geodata provides access to a vast amount of georeferenced imagery 
with city-wide \cite{lin2015learning} or even nation-wide geographic coverage
\cite{vallone2022danish}. We adopt the same idea but use it for a high altitude
aerial-to-aerial scenario.
Most related to ours is the work by Zaffar et al. 
\cite{zaffar2019state}. It investigates an aerial-to-aerial dataset and 
poses the question whether off-the-shelf VPR-techniques 
can cope with the viewpoint variation imposed by the 6-DoF movement of aerial 
platforms compared to the lateral movement of ground platforms. Its scope,
however, is limited to scenarios with low-altitude drone flight at building 
level. In contrast to the previously published VPR datasets, ours centers around 
images taken from a vantage point hundreds of meters above ground with a 
downwardfacing camera which opens up a new research direction for VPR and VL
dealing with challenges induced by the large distance to the scene and in-plane 
rotations. 

\section{The VPaiR Dataset}

\begin{figure}[t]
  \centering
  \centerline{\includegraphics[width=8.5cm]{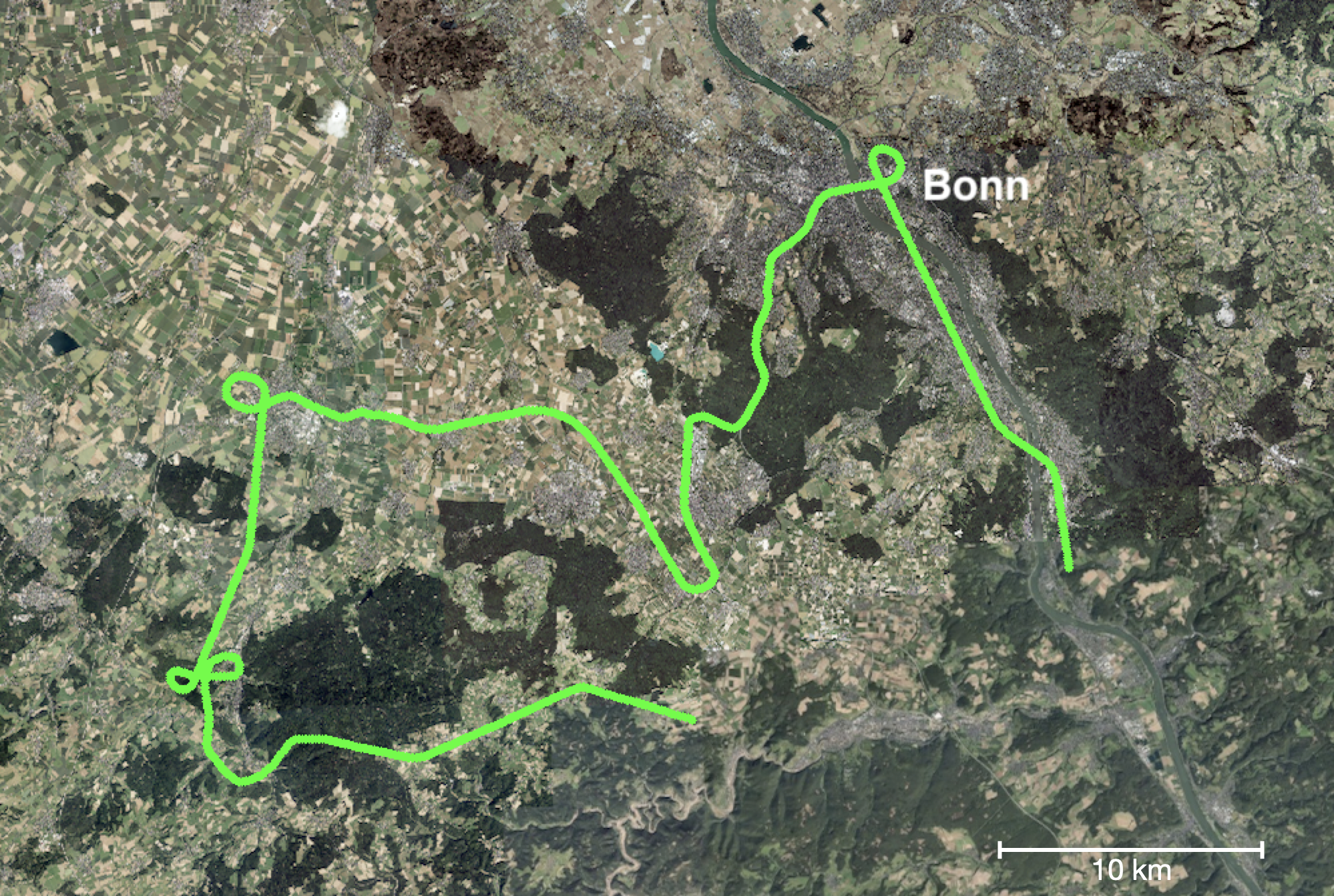}}
\caption[]{{Recording area. The flight trajectory covers various types of landscapes, i.e. urban, rural, forestry and agricultural, and is 
107 km long in total.}}
\label{fig:flight_path}
\end{figure}

VPAIR is a dataset for evaluating visual place recognition and localization in a large-scale 
aerial environment.  
Our goal was to collect data over a wide area of surface 
types to be challenging and representative. The VPAIR dataset was 
recorded with a light aircraft over a 
region between the city of Bonn, Germany, and the Eifel mountain range at 300 to 
400 meters above ground covering a distance of 107 kilometers. For a depiction
of the flight path see Fig. \ref{fig:flight_path}. The dataset includes 
camera images from a downward-facing camera and timestamped reference 
poses recorded with a precise satellite and inertial navigation system 
(GNSS/INS). The sensor data was recorded on October 13, 2020 in a single pass. 
It is complemented with high resolution 3D rendered and geotagged reference 
imagery for the geolocalization task, dense depth maps and metadata that describes the dominant 
surface type visible in the image for evaluation purposes. In total 
there are 2788 query images and 12788 database images. 

\subsection{Sensor Setup}

The data was collected onboard a lightweight 2-seat fixed-wing aircraft 
with a nominal cruise speed of around 150 km/h. The payload consists of 
a monocular color camera recording with 1600x1200 resolution at 25 Hz and 
a dual antenna Ellipse-D GNSS/INS navigation system from SBG which provides 
accurate 6-DoF poses with an expected uncertainty of 0.05$^{\circ}$ in 
rotation and less than one meter in position. The image data is downsampled
to 800x600 pixels and 1 Hz for this dataset
to limit the total file size and increase its practicality while maintaining 
visual coverage of the full trajectory.  

\begin{figure}[t]
    \centering
    \centerline{\includegraphics[width=8.5cm]{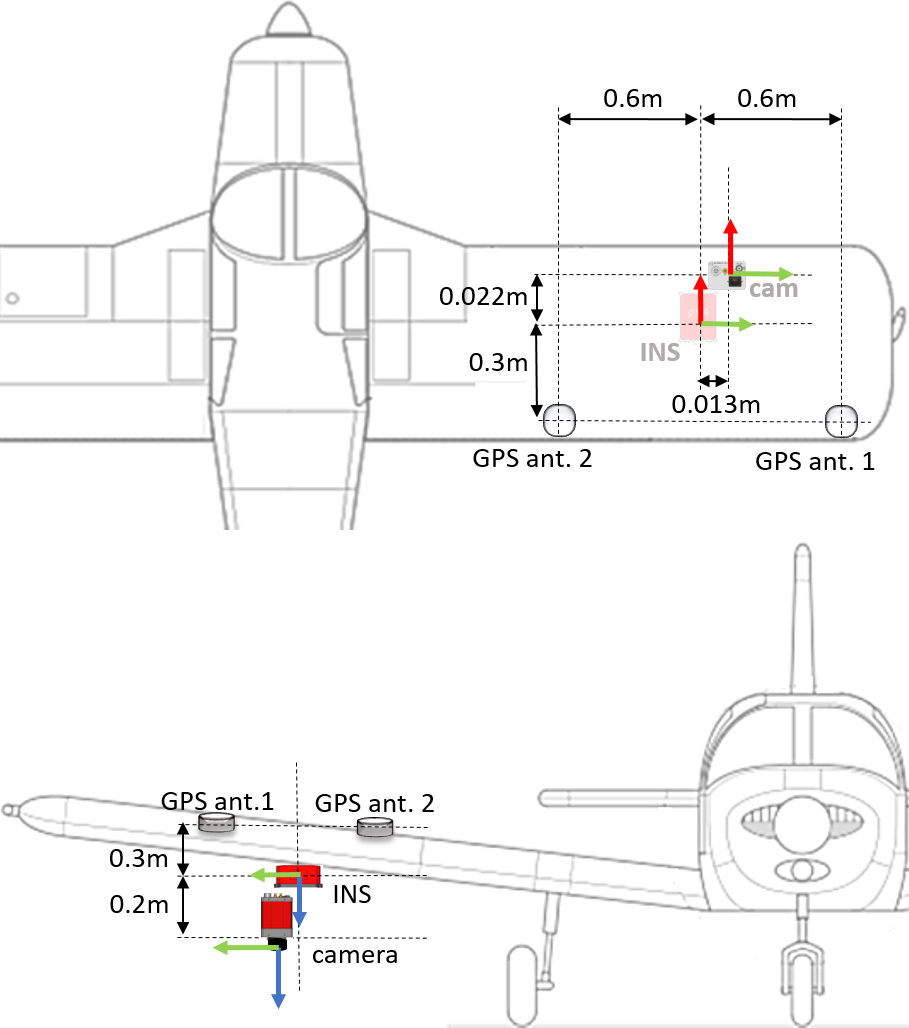}}
  \caption[]{Diagram of the light aircraft (top and front view) and the sensors
    locations.
    The distances between the GPS Antennas and INS have been  
    measured manually while
    the distances between the sensors are obtained via a camera-inertial 
    calibration process. 
    The coordinate frames show the origin and direction of each sensor 
    mounted 
    to the vehicle with the convention: x-forward (red), y-right (green), 
    z-down (blue).}
  \label{fig:payload}
  \end{figure}

All items of the payload were placed close to each 
other below the wing with the camera facing down except for the GNSS 
antennas which were mounted on the top of the wing. We measured the distances 
between all parts of the payload carefully and provide a schematic in Fig. 
\ref{fig:payload}.

The Camera and INS are synchronized 
through hardware triggering ensuring accurate timestamps. The camera's 
timestamps correspond to the beginning of exposure. The exposure time was 
fixed to 5 ms at the day of capture. We used the official ROS 
implementations of the sensor drivers and adapted them for timestamp 
synchronization. The camera intrinsics and the extrinsics between 
IMU and camera were obtained by using the Kalibr calibration toolbox 
\cite{rehder2016extending}.

\subsection{Reference Imagery}

Each image from the onboard camera is paired with a spatially aligned 
reference image, dense metric depth map and metadata describing the dominant 
type of land cover (i.e. urban, agricultural, forestry). 
The reference images and depth maps
are rendered using a self implemented 3D engine in OpenGL in conjunction 
with publicly available orthophoto imagery and 
3D surface models. The latter are provided by Geobasis NRW, a state funded
geodata repository, that is accessible with a permissive open data license 
and covers the complete 
territory of the state of Nordrhein-Westfalen, Germany\footnotemark[1]. 
Land cover information is also obtained via Geobasis. 
The surface model is represented by
a 3D point cloud with approximately 0.5 m accuracy of not only the ground 
but also vegetation
and buildings derived from airborne laser scans. The orthophotos were 
captured between 2019 and 2021 
and are provided with a ground resolution of 0.1 m per pixel. 

\begin{figure*}[t]
  \centering
  \centerline{\includegraphics[width=\textwidth]{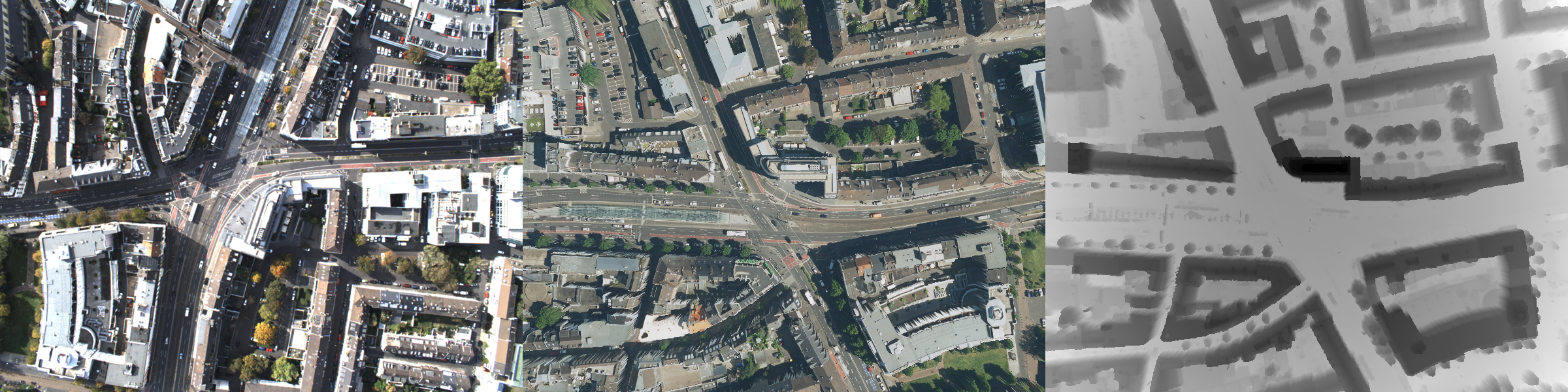}}
\caption[]{{A sample from the dataset. Left to right: query image, reference image and metric depth map.}}
\label{fig:sample}
\end{figure*}

\footnotetext[1]{Data was accessed via https://www.geoportal.nrw and is 
redistributed according to Data license Germany - Zero - Version 2.0 http://www.govdata.de/dl-de/zero-2-0}

The 3D point cloud is processed by the 3D rendering engine to a mesh and 
then textured using the orthophotos. Given 
a 6-DoF pose and an intrinsic camera model the engine is then able to output 
a perspective image that very closely mirrors the original view. 
When creating the reference imagery for this dataset, however, we disregard 
roll, pitch and yaw by setting them to a constant angle resembling a perfectly 
downwardfacing camera where the top of the image is facing in north direction.
In addition to the image pairs we include 10.000 
distractor images, that were sampled from a geographically separate region 
in a grid pattern within an area of 400 km$^2$ near the 
city of Duesseldorf.

\section{Experimental Setup}
Given a query image from an onboard camera the goal of VPR is to recognize 
images of the same place within a large collection of georeferenced images. 
Based on this match we are then provided with a course location 
estimate. Proceeding with a coarse position estimate and information about the
3D scene structure it is then the goal of VL to refine the estimate and provide
a 6-DoF pose. We use the off-the-shelf rotation-robust descriptors (RoRD) from Parihar et al. \cite{parihar2021rord}
for both the VPR and VL task and compare them to NetVLAD and SIFT/D2-Net respectively.

\subsection{Visual Place Recognition}

Each reference image in the dataset is described by a set of local descriptors which are then
all stored in a database. For each
query we retrieve the $k=10$ nearest neighbors of each query's feature descriptors. The 
retrieved features are then associated with their source images. Finally, a ranking 
can be computed based on the occurrence of retrieved features per database image.
The highest ranking reference images constitute the place candidates. 

We compare this VPR pipeline based on local features against a widely 
used off-the-shelf technique NetVLAD \cite{arandjelovic2016netvlad} which 
has exhibited favorable performance in aerial \cite{zaffar2019state} and 
large-scale autonomous navigation scenarios \cite{zaffar2021vpr}
in the past. We use the standard Recall@$n$ metric for evaluating the VPR 
techniques with a distance threshold of 100 m.

\subsection{Visual Localization}

Assuming that the correct image was retrieved the pose is refined 
based on a PnP solver. Given a set of 2D-3D point correspondences it estimates
a 6-DoF pose by minimizing the reprojection error in the camera plane. Our 
VL pipeline works as follows. 
First 2D-2D matches between query and reference image are obtained with 
the rotation robust descriptors. Then, similar to \cite{vallone2022danish} 
we retrieve 3D scene information by back-projecting the rendered image based 
on dense metric depth maps thereby obtaining the 3D global scene coordinates
for each matched keypoint. Finally, the PnP solver provides the 6-DoF pose.
We compare the usage of RoRD against two off-the-shelf baselines, 
namely SIFT and D2-Net evaluating on the absolute pose error metric \cite{sattler2018benchmarking}.

\section{Results}

Table \ref{tab:vpr} and \ref{tab:vpr_landcover} show that the VPR-technique 
based on off-the-shelf local descriptors compares favorably to the global image 
descriptor NetVLAD. Performance is generally better in scenes that depict urban 
environments which might be less subject to appearance change than scenes depicting agricultural environments or forests. 
Table \ref{tab:vpr_angle} compares the place recognition performance when 
controlling for the difference in heading. 
NetVLAD suffers drastically from in-plane rotation while RoRD-VPR's 
performance drop is less significant. 

We note that our VPR pipeline based on 
local features requires hundreds of retrieval operations per image compared 
to a single one when using global image descriptors like NetVLAD.
The runtime and memory requirements are therefore impractical with a large 
database especially on constrained mobile platforms. 
This highlights the need for rotation robust global image descriptors.

Finally, Table \ref{tab:visloc} shows that RoRD outperforms
D2-Net and SIFT in the visual localization task. D2-Net is a learned image 
descriptor that is based on the same architecture but not rotation robust. 
SIFT is rotation-invariant but suffers
from the appearance change between query and reference images. 

\begin{table}[]
  \centering
  \caption{VPR Evaluation based on Recall@n.}
  \begin{tabular}{@{}llll@{}}
  \toprule
           & R@1  & R@5  & R@20 \\ \midrule
  NetVLAD  & 10.2 & 22.6 & 35.7 \\
  RoRD-VPR & \textbf{38.7} & \textbf{50.1} & \textbf{58.8}
  \end{tabular}
 
  \label{tab:vpr}
  \end{table}
\begin{table}[]
  \caption{Recall@5 grouped by dominant scene surface type}
  \centering
  \begin{tabular}{llll}
  \hline
            & urban & agricultural & forestry \\ \hline
  NetVLAD  & 31.1  & 15.6         & 5.5      \\
  RoRD-VPR & \textbf{77.4}  & \textbf{27.4}         & \textbf{16.9}    
  \end{tabular}
    
    \label{tab:vpr_landcover}
    \end{table}
    \begin{table}[]
      \caption{Effect of varying difference in heading between query and reference images on Recall@5. Without distractors. }

      \centering
      \begin{tabular}{lllll}
      \hline
      Heading difference  & 0$^{\circ}$ & 30$^{\circ}$& 90$^{\circ}$ &135$^{\circ}$ \\ \hline
      NetVLAD  & 66.1        & 34.4 & 23.3 & 17.7         \\
      RoRD-VPR & \textbf{71.4} & \textbf{69.4} & \textbf{56.9}& \textbf{51.2}         
      \end{tabular}
      \label{tab:vpr_angle}
      \end{table}

\begin{table}[]
  \centering
  \caption{Ratio that satisfies both translational and rotational thresholds measured in absolute pose error.}
  \begin{tabular}{lll}
  \hline
          & 25m/5$^{\circ}$ & 50m/10$^{\circ}$ \\ \hline
  SIFT    & 6.2             & 15.9             \\
  D2-Net  & 2.9             & 7.0              \\
  RoRD-VL & \textbf{10.6}            & \textbf{27.8}            
  \end{tabular}

  \label{tab:visloc}
  \end{table}

\section{Conclusions}

We propose VPAIR - a challenging dataset for aerial visual place recognition and localization.
Experiments based on off-the-shelf image descriptors highlight the need for rotation robust and 
runtime efficient VPR and VL techniques. By releasing the dataset to the community we hope to 
foster research in large-scale aerial geolocalisation.

\addtolength{\textheight}{-12cm}   




\bibliographystyle{IEEEtran}
\bibliography{IEEEabrv,refs}

\end{document}